\def\year{2020}\relax
\documentclass[letterpaper]{article} % DO NOT CHANGE THIS
\usepackage{amsmath}
\usepackage{aaai20}  % DO NOT CHANGE THIS
\usepackage{times}  % DO NOT CHANGE THIS
\usepackage{helvet} % DO NOT CHANGE THIS
\usepackage{courier}  % DO NOT CHANGE THIS
\usepackage[hyphens]{url}  % DO NOT CHANGE THIS
\usepackage{graphicx} % DO NOT CHANGE THIS
\graphicspath{ {} }
\usepackage{subcaption}
\newcommand{\add}[1]{{#1}}
\newcommand{\del}[1]{\ignorespaces}
\newcommand{\rem}[1]{\ignorespaces}
\newcommand{\madd}[1]{{#1}}
\newcommand{\mdel}[1]{\ignorespaces}
\newcommand{\com}[1]{\ignorespaces}
\newcommand{\mcom}[1]{\ignorespaces}
\newcommand{\eqdel}[1]{\ignorespaces}
\usepackage{xcolor}

\title{Unsupervised Embedding Learning for Human Activity Recognition Using Wearable Sensor Data}
\author{\Large \textbf{Taoran Sheng and Manfred Huber} \\ 
Department of Computer Science and Engineering\\ %If you have multiple authors and multiple affiliations
University of Texas at Arlington\\
taoran.sheng@mavs.uta.edu, huber@cse.uta.edu % email address must be in roman text type, not monospace or sans serif
}
 
\begin{document}

\maketitle

\begin{abstract}

The embedded sensors in widely used smartphones and other wearable devices make \del{collecting} the data of human activities more accessible\del{, h}\add{. H}owever, recognizing different human activities from the wearable sensor data remains a challenging research problem in ubiquitous computing. One of the reasons is that the majority of the acquired data has no labels. In this paper, we present an unsupervised approach, which is based on the nature of human activity, to project the human activities into an embedding space \mdel{,} in which similar activities will be located closely \madd{together}. \del{Thus}\add{Using this,} subsequent clustering algorithm\add{s} can \del{be} benefit from the embeddings\add{, forming behavior clusters that represent the distinct activities performed by a person}. \del{The r}\add{R}esults of \del{the} experiments on three \add{labeled} benchmark datasets \del{have} demonstrate \del{d} the effectiveness of the framework and show \del{n} that our approach can help the clustering algorithm achieve improved performance \add{in identifying and categorizing the underlying human activities compared to unsupervised techniques applied \madd{directly} to the original data set}\del{ in the task}.

\end{abstract}

\section{Introduction}
\begin{figure}
\begin{center}
\includegraphics[width=0.95\columnwidth]{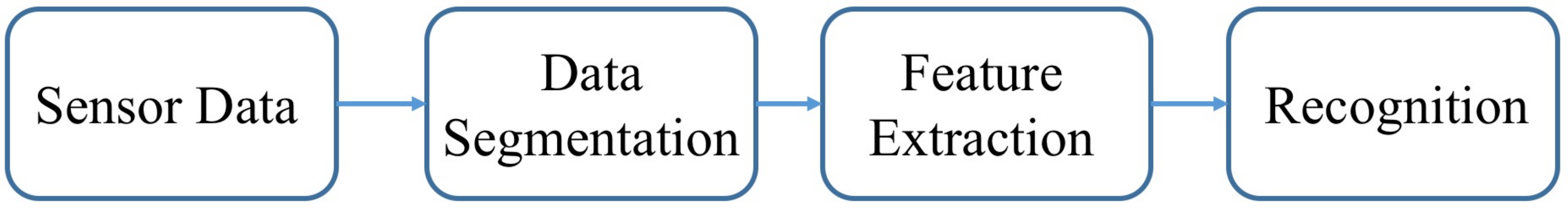}
\caption{The typical process of HAR.} \label{arc}
\end{center}
\end{figure}

The typical process of sensor-based human activity recognition (HAR), as shown in Figure\mdel{.} \ref{arc}, consists of three important stages: data segmentation, feature extraction, and recogniz\add{ing} the type of the activity. Extensive studies have been conducted in all the stages of the HAR process\del{\cite{BullingTutor}}. However, existing HAR methods rely heavily on \del{the} labeled data to supervise the model training and \add{to} perform the recognition\del{\cite{DBNHAR}}, thus a huge challenge for HAR system\madd{s} is collecting annotated data. In the meantime, more and more wearable devices, smartphones, smart watches, etc., are used in people's daily lives. These wearable devices are usually equipped with various sensors, such as accelerometers, gyroscope, GPS sensors\madd{,} etc., which can provide \add{a} massive amount of unlabeled sensor activity data. Due to the above mentioned facts, most of the existing HAR systems can not take advantage of the accessible unlabeled data efficiently. Therefore, we aim at developing an unsupervised method to leverage the unlabelled data to recognize the physical activities without requiring \mdel{the} labels.

\add{The} Autoencoder (AE) \del{\cite{vanillaAE}}\com{We need a reference here}is an unsupervised learning framework \del{for}\add{to find} efficient encoding\add{s of} data. It encodes important features of the inputs into \del{the}\add{a hidden} representation\del{s}, and then reconstructs the inputs based on the\add{ir hidden} representations. It is applied for dimension reduction, deep hierarchical model pre-training\madd{,} etc. Because of its simplicity and efficiency, we also design our model based on the AE architecture. Yet only using reconstruction to guide the learning can \del{make}\add{lead} the AE \add{to} encode \del{much}\add{a significant amount of} unnecessary information, such as task-irrelevant information, or even destructive information, such as noise.

Motivated by this observation, our approach utilizes the intrinsic properties of the sensor activity data to project the data into a clustering-friendly embedding space. \add{Two fundamental observations contribute to the design and formation of this space.}

First\add{ly}, slow feature analysis \cite{slowfeature} finds that many properties in the real world change\del{s} slowly over time. Physical objects have inertia and their states usually change gradually and infrequently. This rule \add{also} applies to human activity. In most \del{of the} situations, the period of a person performing an activity will take a \del{bit}\add{a relatively significant amount of time}. It is rare that a person will switch between different activities very frequently. \del{And}\add{On the other hand, while} during the course of an activity\del{, while} the type of the activity remains the same, the body pose of the activity varies over \mdel{the} time. Hence, we include the temporal coherence property in the loss function, which \del{let}\add{allows} the model \add{to} learn the essential features of the activity and ignore the irrelevant temporal details in \add{the} body pose.

Second\add{ly}, distinct activity and person characteristics are commonly co-present in the sensor data. For the HAR task, only the activity characteristics are relevant. For example, two persons can walk in two different styles, but people can still identify they are performing the same type of activity: walking, because the activity-relevant characteristics are used in the recognition process and the irrelevant person characteristics are disregard. Based on this property, another local neighborhood based objective function, which aims at removing irrelevant personal or individual details in the data, is used to guide the learning of the model.

\section{Related Work}
Many works have been proposed to recognize human activity with wearable sensor data. As illustrated in Figure. \ref{arc}, the first step in HAR is \madd{generally} to segment the sensor data sequence. One common method used to \madd{do this}\mdel{segment the sequence} is \del{the}\add{to use a} sliding window\del{\cite{windowSegmentation}}\com{is there a reference we could include regarding segmentation of time series ?}. We also adopted this method in our model for its simplicity and tractability. Feature extraction and recognition are conducted on the segmented data. To recognize the activity type, discriminative features are needed. They can be designed with domain knowledge or extracted automatically \del{by}\add{using, for example,} neural network\add{s} (NN). % \cite{Plotz2011FL}. 

Handcrafted features, when designed properly, \del{are proved}\add{have proven} to be very useful in HAR systems. In \cite{wisdm}, statistical features are derived from the time series sensor data. In \cite{dctFeature}\del{and \cite{waveletFeature}}, discrete cosine transform \del{and wavelet transform are} is used to convert the sensor signal from \add{the} time domain into \add{the} transformed domain. Then features derived from \add{the} transformed domain\mdel{s} are applied in the recognition process. 

With the development of \add{more competent deep learning techniques, and in particular} NN\add{s}, automatic feature extraction \del{is}\add{has become} another effective way to obtain discriminative features. In \cite{DBNHAR}, a deep belief network was used as the emission matrix of a hidden Markov model. In \cite{cnnLSTMHAR}\del{and \cite{cnnHAR}}, convolutional neural networks \del{and recurrent neural networks are} \mdel{is}\madd{are} employed to extract the features and recognize the type of the activity.

\del{But}\mdel{\add{However}, t}\madd{T}he features derived by these methods are \del{all}\add{then} applied in \del{the}\add{a} subsequent supervised model \del{s} \add{learning phase} \del{\cite{unsupervisedFL}}\del{. B} \add{b}ecause the\del{se features}\add{y} are usually not \add{sufficient}\del{good enough} for direct\del{ly} us\del{ing}\add{e} in an unsupervised model. There are \add{only} a few works that are recognizing the activity type \del{via}\add{in} an unsupervised manner.\del{A variational AE is used in \cite{motion2vec} to compress the sensor data into meaningful features.} In \cite{Lu2017}, a protein interaction method was used to cluster the activity data. In \cite{Kwon2014UnsupervisedLF}, a set of time-frequency domain features are adopted, and unsupervised methods\del{:}\add{, in particular} DBSCAN\del{,} \add{and} mixture of Gaussian are used to cluster five basic activities.
\com{Can we say something regarding the shortcomings of previous work ?}

\section{Approach}

\begin{figure}
\begin{center}
\includegraphics[width=0.95\columnwidth]{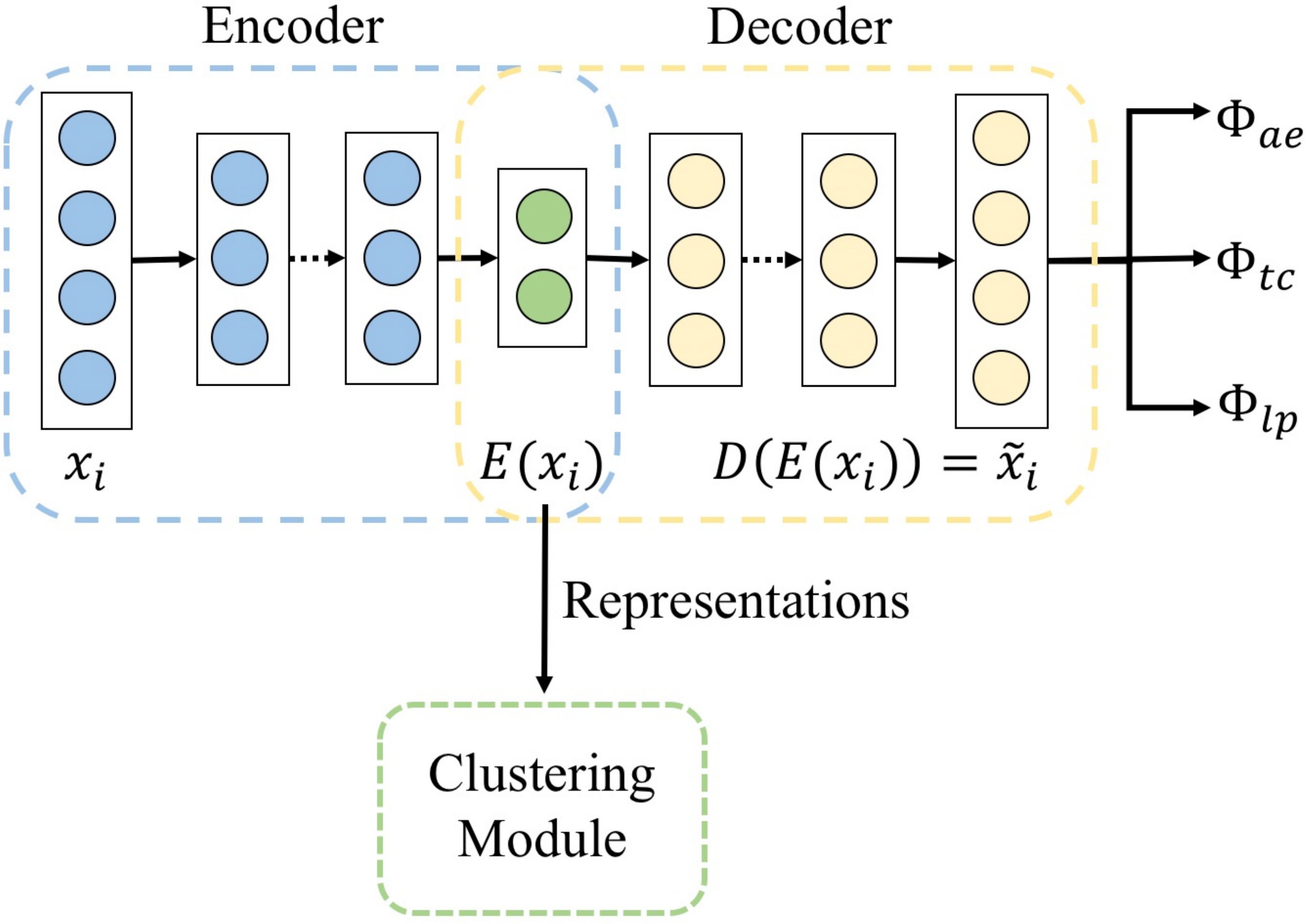}
\caption{The overall architecture of the approach.} \label{structure}
\end{center}
\end{figure}

Our approach differs from other works by using the \del{knowledge}\add{aforementioned properties} with regard to the nature of the activit\del{y}\add{ies}. Specifically, our approach attempts to leverage two types of relationships: the temporal coherence of time series data, and \del{the} locality preserv\del{ing}\add{ation} in the feature space.

\subsection{Architecture}
As shown in Fig\mdel{.}\madd{ure~}\ref{structure}, the foundation of the overall architecture of our approach is an AE framework. It consists of two parts: an encoder, and a decoder. The encoder defines the transformation: $E(\cdot)$, which transforms the input data sample $x_i$ to the representation $E(x_i)$; and the decoder defines another transformation: $D(\cdot)$, which \add{attempts to} reconstruct \del{s} the original input $x_i$ based on the its representation $E(x_i)$:
\[
\tilde{x_i} = D(E(x_i))
\]
where, $\tilde{x_i}$ is the reconstructed input. In the traditional AE, the loss function of one data sample can be defined as:
\[
\Phi_{ae}(x_{i}) = ||x_{i} - \tilde{x_{i}}||^2
\]
where $||\cdot||$ denotes the Euclidean distance. This loss function forces the reconstruction $\tilde{x_i}$ to be as similar as possible to the original input $x_i$. If a good reconstruction $\tilde{x_i}$ can be decoded from the representation $E(x_i)$, it means \madd{that} the representation $E(x_i)$ has retained much of the information that is important in the input $x_i$, so that the reconstruction $\tilde{x_i}$ can be very similar to the original input $x_i$. \del{And}\add{Thus} the representation $E(x_i)$ can be used in other tasks, such as \del{,} classification or clustering.

However, merely retaining information for reconstruction is usually not enough. The aim of the traditional AE is to learn a representation $E(x_i)$ that contains sufficient information to reconstruct the input, so an exact reconstruction also means to reconstruct noise and all the details in the input data. But not all the information in the learned representation is relevant to the subsequent task (e.g. noise \del{,} \add{as well as} some task-irrelevant details \add{might not only be unnecessary but could even be detrimental, especially in the context of subsequent clustering}). Therefore, more task-oriented loss functions are imposed in our approach to guide the learning of the AE and make the learned representations more useful in the subsequent clustering task.

\subsection{Temporal Coherence}
Intuitively, \del{the}\add{a} human activity can be decomposed into two components, a temporal\add{ly} varying component and a temporal\add{ly} stationary component. Specifically, certain dynamic properties of a\del{n} \add{single} activity can vary over the time. For example, while walking \del{,} the body pose varies over \del{the} time: left foot and right foot alternatively step forward. This \del{kind}\add{type} of dynamic property is recorded in the sensor data\madd{,} too\add{, and w}\del{. W}e refer \madd{to} it \add{here} as the temporal\add{ly} varying component. On the other hand, no matter how the body pose varies over \mdel{the} time, the semantic content of the activity remains the same. Namely, left foot and right foot can step forward alternatively, but the type of the activity is still walking. We refer \madd{to} this part as the temporal\add{ly} stationary component. 

Based on this nature of the human activity, the temporal coherence loss forces temporally close data samples to be similar to one another, and ignore the difference in the temporal varying component. It is motivated by the intention that the semantic content\del{:}\add{, i.e.} the type of the activity, in which we are interested, should vary \add{relatively} infrequently over \del{the} time. If the data samples are temporally close to each other, they may represent the same type of activity, even \add{as} they may be very distant in terms of the Euclidean distance in the sensor data space. The temporal coherence loss preserves the temporal continuity of the sensor data.

More formally, let $x_{i}^{t}$ denote a data sample $i$, which occurs at time \del{point} $t$ \del{over}\add{during} the course of an activity. Let $M_{i}^{t}$ denote the ind\del{ices}\add{ex} set of $m$ temporal neighbors\add{, $x_{j}$,} of $x_{i}^{t}$\del{, $x_{j}$ an element in the set $M_{i}^{t}$}. Then the temporal coherence loss $\Phi_{tc}$ for $x_{i}^{t}$ is defined as:
\[
\Phi_{tc}(x_{i}^{t}) = \frac{1}{m}\sum_{j \in M_{i}^{t}}||x_{j} - \tilde{x_{i}^{t}}||^{2}
\]

The temporal coherence loss encourages the reconstruction $\tilde{x_{i}^{t}}$ to be similar to its temporal neighbors so that the encoder can extract useful features from \add{the} temporal\add{ly} stationary component and ignore irrelevant time varying details.

\subsection{Locality Preserv\del{ing}\add{ation}}
\add{Locality preservation is i}\del{I}nspired by the observation that different persons perform the same type of activity in different fashions, but different fashions don't hinder other people to identify the activity type. Hence we assume that the personal or individual features in the activity data may not be necessary in the activity clustering stage, and the features \mdel{,} which are commonly present across multiple data points \mdel{,} may be the essential features of the activity. The locality preserving loss function is based on this assumption. 

In \del{the} past research works, the combination of \del{the} carefully designed handcrafted \add{high level} features \add{to represent the main characteristics of a temporally varying signal value, } and \add{of} the k-Nearest Neighbor algorithm \del{is} proved to be a powerful method to classify the sensor data of \del{the} human activit\del{y}\add{ies}\del{\cite{pamap2}}\com{We need a reference here, even if it is a repetition.}. Due to its effectiveness and simplicity, it is employed in this approach to define the local neighborhood of a data sample. \del{And t}\add{T}he locality preserving loss \add{then} aims to preserve the \add{high level} feature \del{s} \add{characteristics} that are generally \del{existed}\add{present} in the local neighborhood. 

The locality preserving loss \del{makes}\add{forces} the decoder to decode a data sample by using the learned representation of its nearby data samples. \add{The rationale here is that}\del{ Because} if the data samples are close to each other in the handcrafted feature space, they may represent the same type of activity. \del{And}\add{Thus} the features shared across multiple nearby data samples should be the essential features of that type of activity. If the features do not exist in all the nearby data samples, then the features may represent personal or individual features, but not activity features. 

Formally, let $x_{i}^{f}$ denote \del{the} data sample $i$ in the feature space, and $\tilde{x_{i}^{f}}$ the reconstruction of $x_{i}^{f}$. Let $N_{i}^{f}$ denote the ind\del{ices}\add{ex} set of $n$ local neighbors\add{,  $x_{k}$, } of $x_{i}^{f}$ in the handcrafted feature space\del{, $x_{k}$ an element in the set $N_{i}^{f}$}. Then the locality preserving loss $\Phi_{lp}$ for $x_{i}^{f}$ is defined as:\com{Is this a notation overload ? You used $f$ for the base features in the AE. Are you using them here for your statistical features ?}

\[
\Phi_{lp}(x_{i}^{f}) = \frac{1}{n}\sum_{k \in N_{i}^{f}}||x_{k} - \tilde{x_{i}^{f}}||^{2}
\]

The locality preserving loss forces the model to recover \del{the} data sample $x_{k}$ with the representation of its nearby point $x_{i}^{f}$. It drives the encoder to encode the information that \del{is}\add{has} generally occurred across the neighborhood and \add{to} disregard individual features of \del{each} single samples.\com{I am not sure if this is clear to the reader. The use of the two feature spaces is not clear here and it is not clear how you use the features you had in the table.}

\subsection{Joint Loss Function}
The joint loss function is the sum of the temporal coherence loss and the locality preserving loss. It is used to train the model and is defined as follows:
\[
min\sum_{i=1}^{S}(1-\alpha-\beta)\Phi_{ae}(x_i) + \alpha\Phi_{tc}(x_{i}^{t}) + \beta\Phi_{lp}(x_{i}^{f})
\]
where $i$ is the index of the sample, $S$ is the size of the dataset, $\alpha$ and $\beta$ are the parameters to balance the contribution of $\Phi_{ae}$, $\Phi_{tc}$, and $\Phi_{lp}$. While $\Phi_{tc}$ and $\Phi_{lp}$ preserve more task relevant information in the representation, the $\Phi_{ae}$ \add{component} is also necessary in the learning process \del{. B} \add{b}ecause without the reconstruction loss $\Phi_{ae}$, the risk of learning trivial solutions or worse representations will \mdel{be} increase \mdel{d} \cite{taxonomy}.

\subsection{Feature Extraction}
After the description of the proposed model, this section focuses on the features used in the experiments. In the feature extraction stage, the segmented raw sensor signals are converted into \mdel{the} feature vectors. Formally, let $r_i$ denote the sample $i$ in the set of the segmented raw sensor signals, $x_i$ the converted feature vector, \madd{and} $C$ the feature extraction function. Then the feature extraction can be defined as:
\[
x_i = C(r_i)
\]
\mdel{The} $x_i$ is used as the input to the proposed model. Table \ref{statisticalFeat} illustrates the \add{statistical high level} features that are used in the proposed approach. Mean, variance, standard deviation, \madd{and} median, which are the most commonly adopted features in the HAR research works, are used in the approach. In addition, some other features, which have been shown to be efficient in previous works \cite{transitionAware}, are included here as well. For example, the feature interquartile range ($iqr$), Quartiles ($Q_1$, $Q_2$ and $Q_3$) divide the time series signal into quarters. \mdel{And}\madd{Using this,} $iqr$ is the measure of variability between the upper and lower quartiles, $iqr = Q_3 - Q_1$. %the feature pearson correlation coefficient between each pair of axes is successfully applied for differentiating activities that include movement in only one dimension or multiple dimensions. 

All these features are computed for each axis separately. Since the data from different sensors is synchronized, combining different sensor data is achievable. In the training process, the \mdel{proposed} model takes these derived features as input and learns to retain the task-relevant information in the features\mdel{,} and to disregard the unnecessary task-irrelevant parts.

\begin{table}[t]
\centering
\caption{List of the used statistical features.}\label{statisticalFeat}\smallskip
\scalebox{0.9}{%
\begin{tabular}{c|c}
\hline
\textbf{Feature extraction function} &  \textbf{Description} \\
\hline
$mean(r_i) = \frac{1}{N}\sum_{j=1}^{N} r_{i_j} $ & Mean \\
\hline
$var(r_i) = \frac{1}{N}\sum_{j=1}^{N} (r_{i_j} - mean(r_i))^2 $ & Variance \\
\hline
$std(r_i) = \sqrt{var(r_i)} $  & Standard deviation \\
\hline
$median(r_i)$ & Median values\\
\hline
$max(r_i)$ & Largest values in array \\
\hline
$min(r_i)$ & Smallest value in array \\
\hline
$iqr(r_i) = Q_{3}(r_i) - Q_{1}(r_i) $ & Interquartile range \\
\hline
%$correlation(s_{1} , s_{2} ) = \frac{cov(s_{1}, s_{2})}{std(s_{1}) std(s_{2})}$ & Pearson correlation coefficient \\
%\hline
\end{tabular}}
\end{table}

\subsection{Cluster Construction}
To train the network, the standard backpropagation algorithm with stochastic gradient descent is used. All the weights are initialized to small values\del{. A} \add{a}nd the network \del{was}\add{is} evaluated on the validation data after each epoch. \del{If}\add{When} the validation error stop\del{ped}\add{s} decreasing for a predefined number of epochs, the training process is complete.

After the learning process \add{to establish the hidden representation within the AE architecture}, the trained model \add{(i.e. the encoder of the architecture)} can project the input data sample $x$ into a clustering-friendly embedding space. More specifically, with the temporal coherence loss and the locality preserving loss, the encoder in the model is learned to encode the essential features across multiple data samples and disregard individual or temporal details that are irrelevant to the clustering task. The learned representations are evaluated in the subsequent clustering task. $k$-means (KM), which is arguably the most popular clustering algorithm, is used in the experiments.

\section{Evaluation and Experiments}
\begin{figure*}
    %\centering
    \begin{subfigure}{0.3\linewidth}        %% or \columnwidth
        \includegraphics[width=1.05\linewidth]{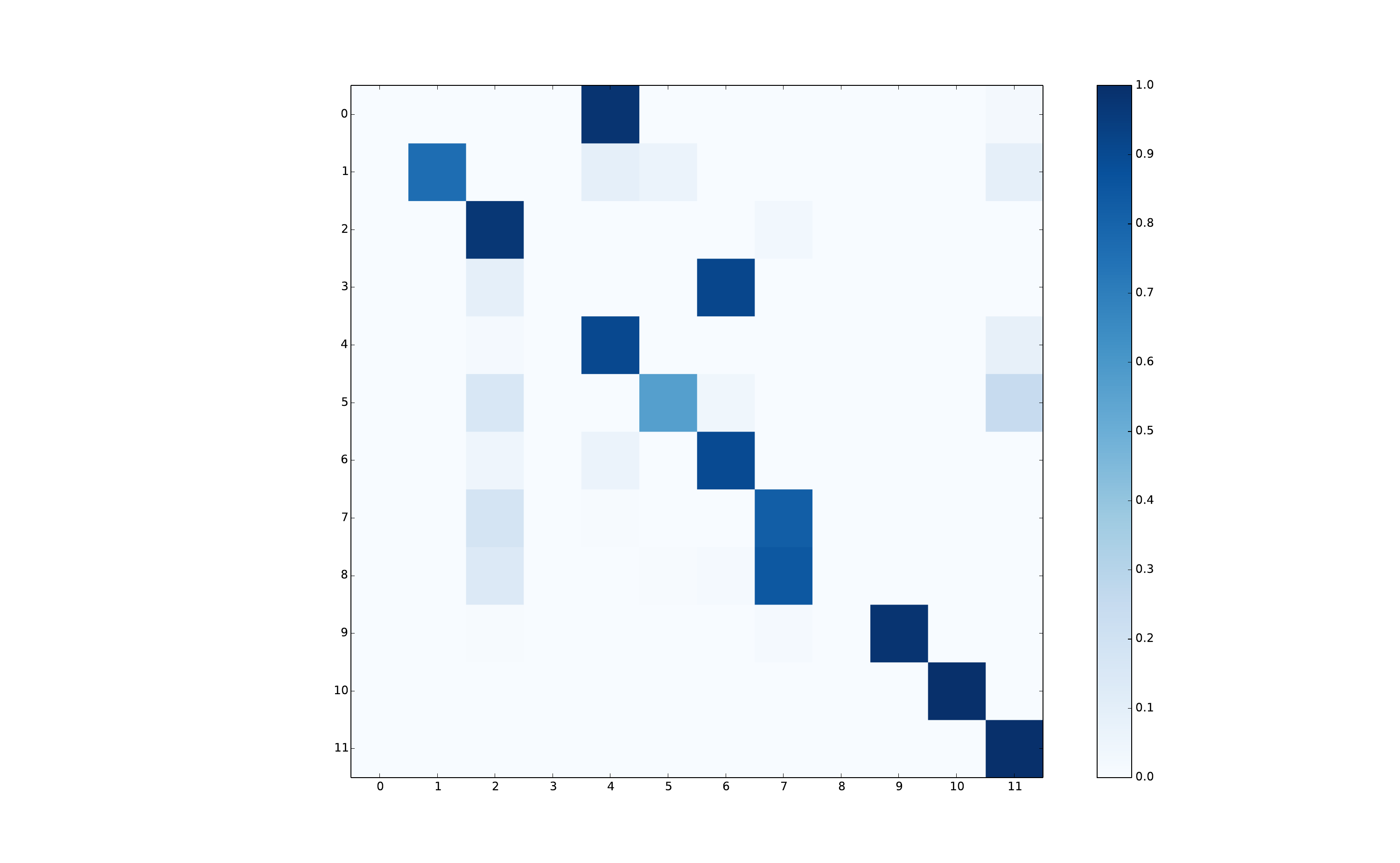}
        \caption{PAMAP2}
        \label{pamap2Result}
    \end{subfigure}
    \begin{subfigure}{0.3\linewidth}        %% or \columnwidth
        \includegraphics[width=1.05\linewidth]{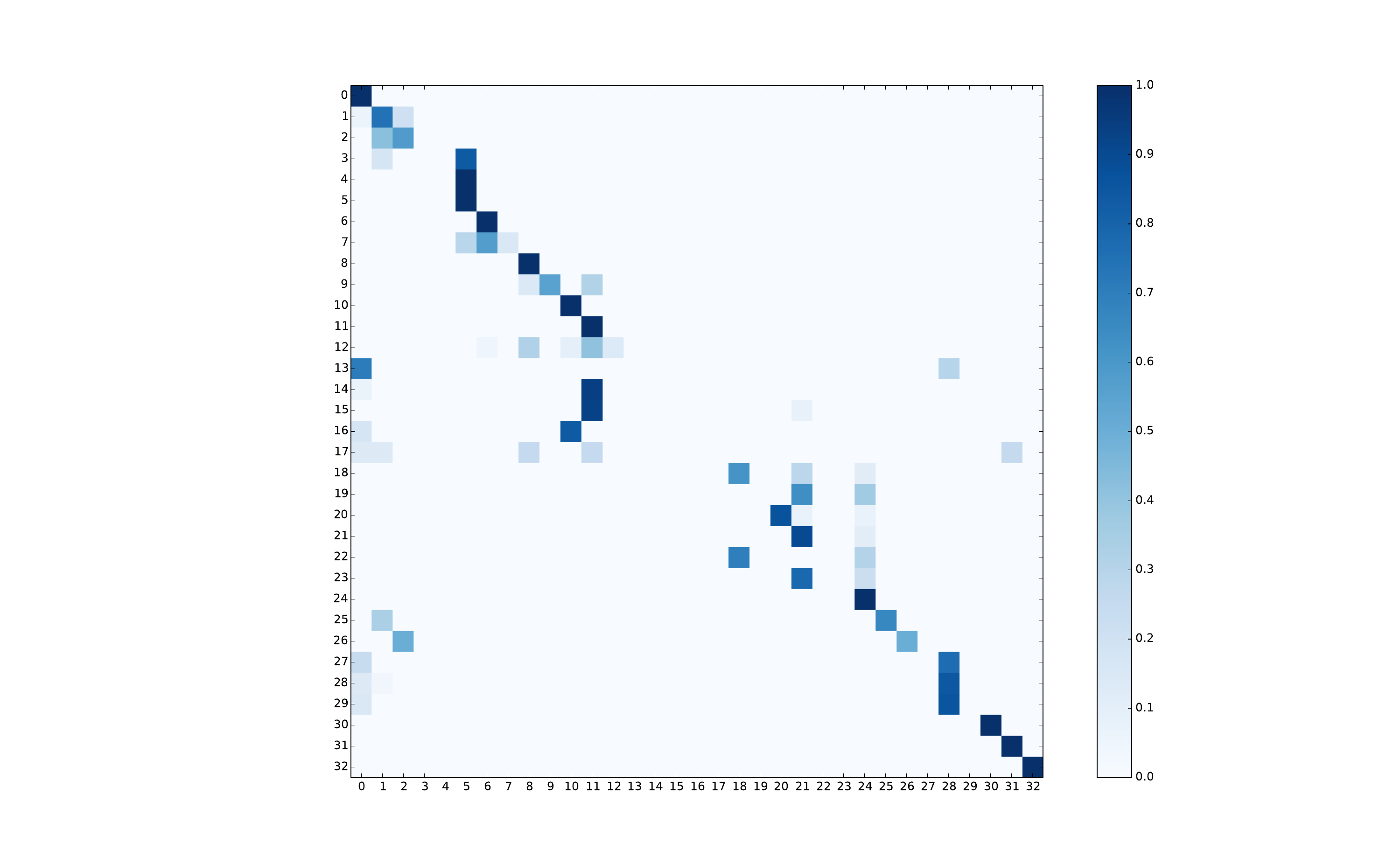}
        \caption{REALDISP}
        \label{realdispResult}
    \end{subfigure}
    \begin{subfigure}{0.3\linewidth}        %% or \columnwidth
        \includegraphics[width=1.05\linewidth]{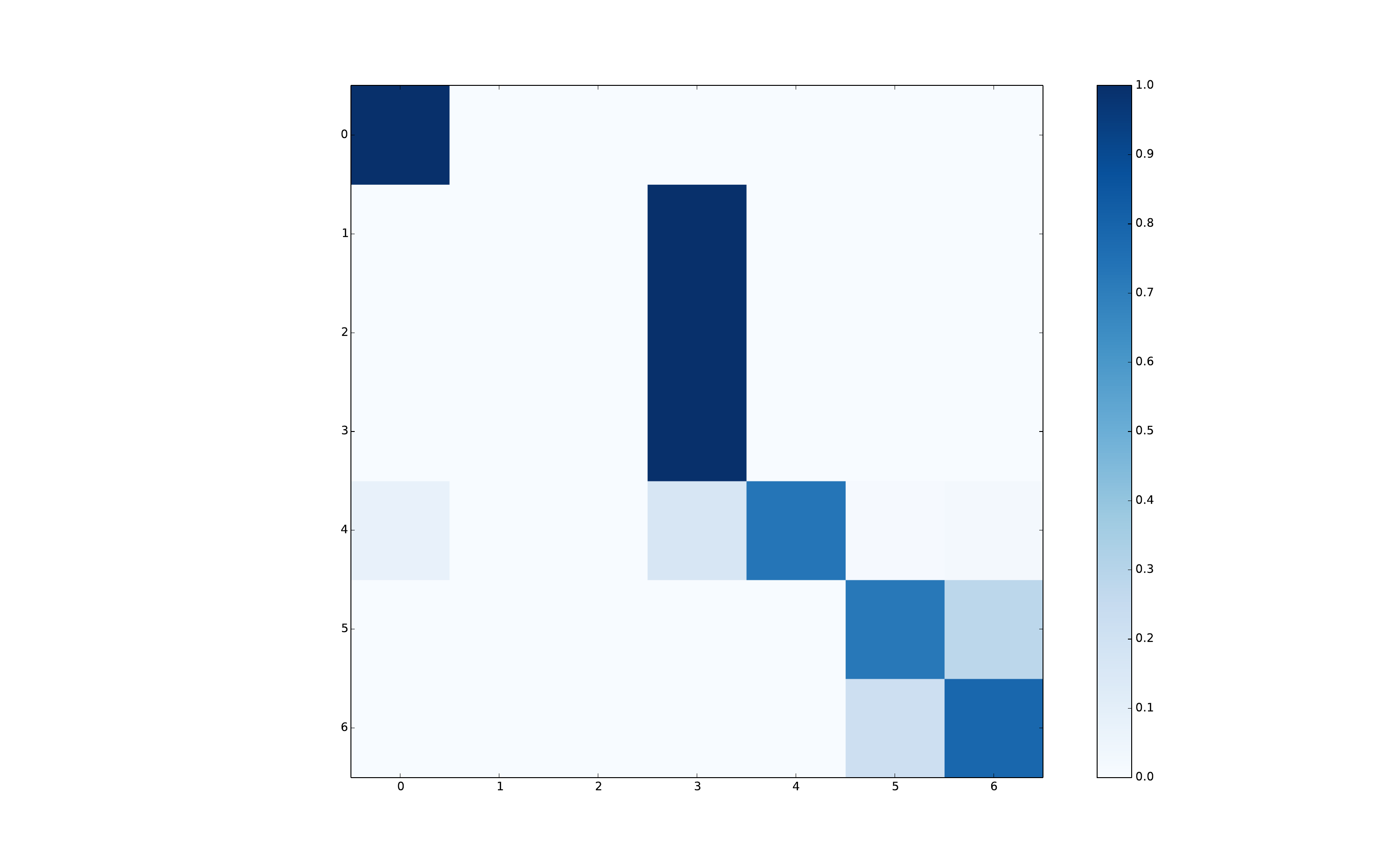}
        \caption{SBHAR}
        \label{sbharResult}
    \end{subfigure}
    \caption{Confusion Matrices of Traditional AE on PAMAP2, REALDISP, and SBHAR.}\label{confusionAE}
\end{figure*}
\begin{figure*}
    %\centering
    \begin{subfigure}{0.3\linewidth}        %% or \columnwidth
        \includegraphics[width=1.05\linewidth]{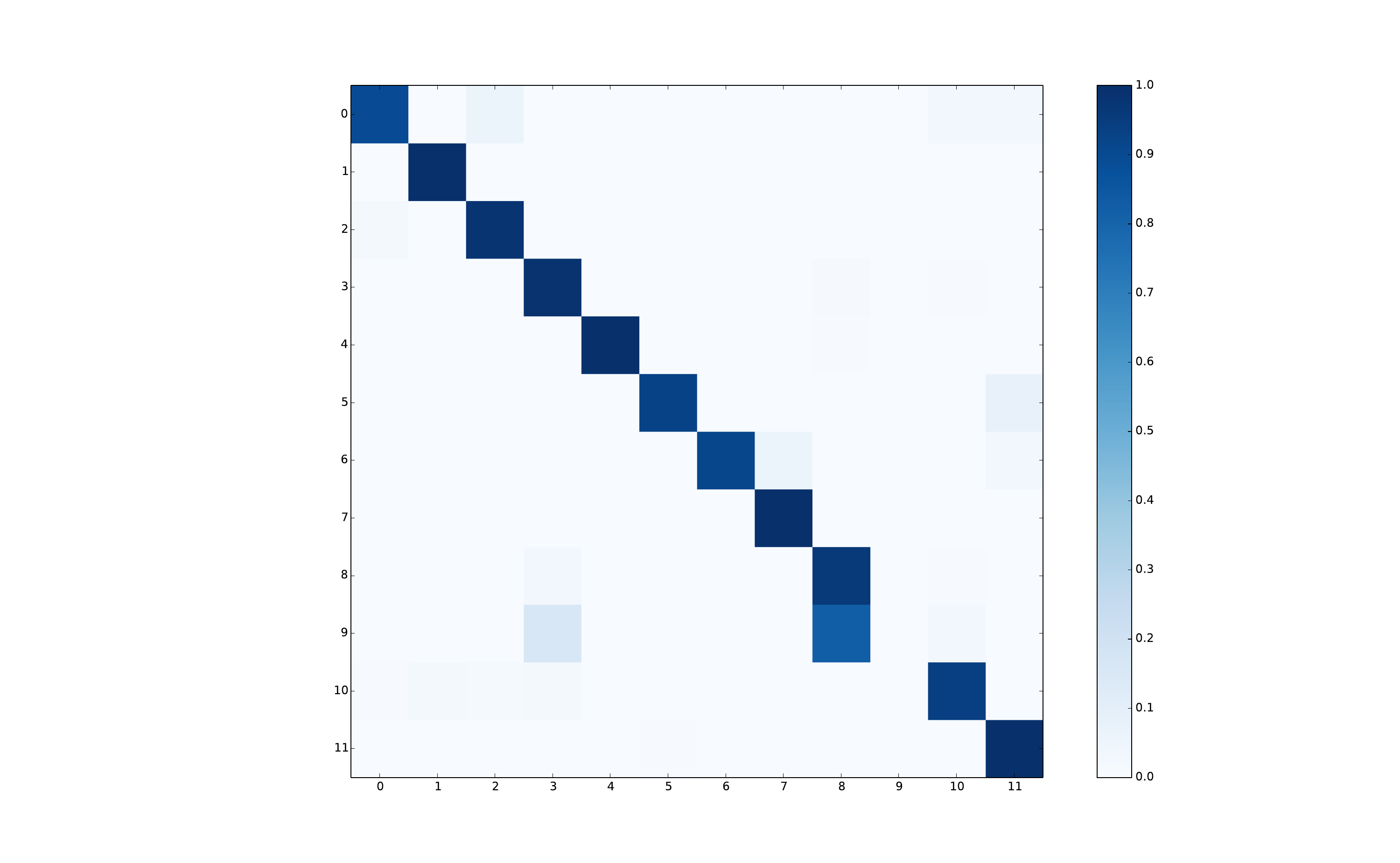}
        \caption{PAMAP2}
        \label{pamap2Result}
    \end{subfigure}
    \begin{subfigure}{0.3\linewidth}        %% or \columnwidth
        \includegraphics[width=1.05\linewidth]{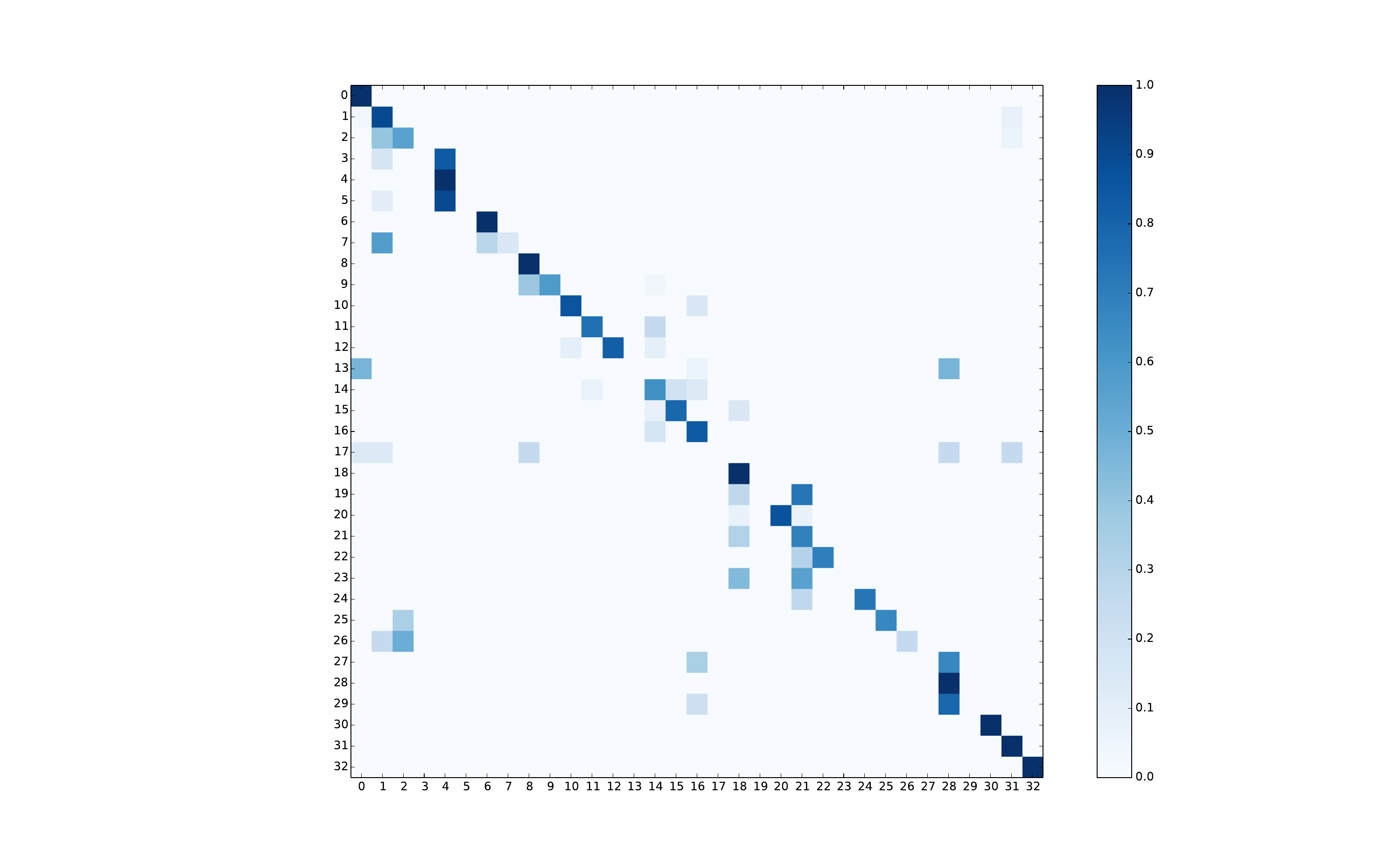}
        \caption{REALDISP}
        \label{realdispResult}
    \end{subfigure}
    \begin{subfigure}{0.3\linewidth}        %% or \columnwidth
        \includegraphics[width=1.05\linewidth]{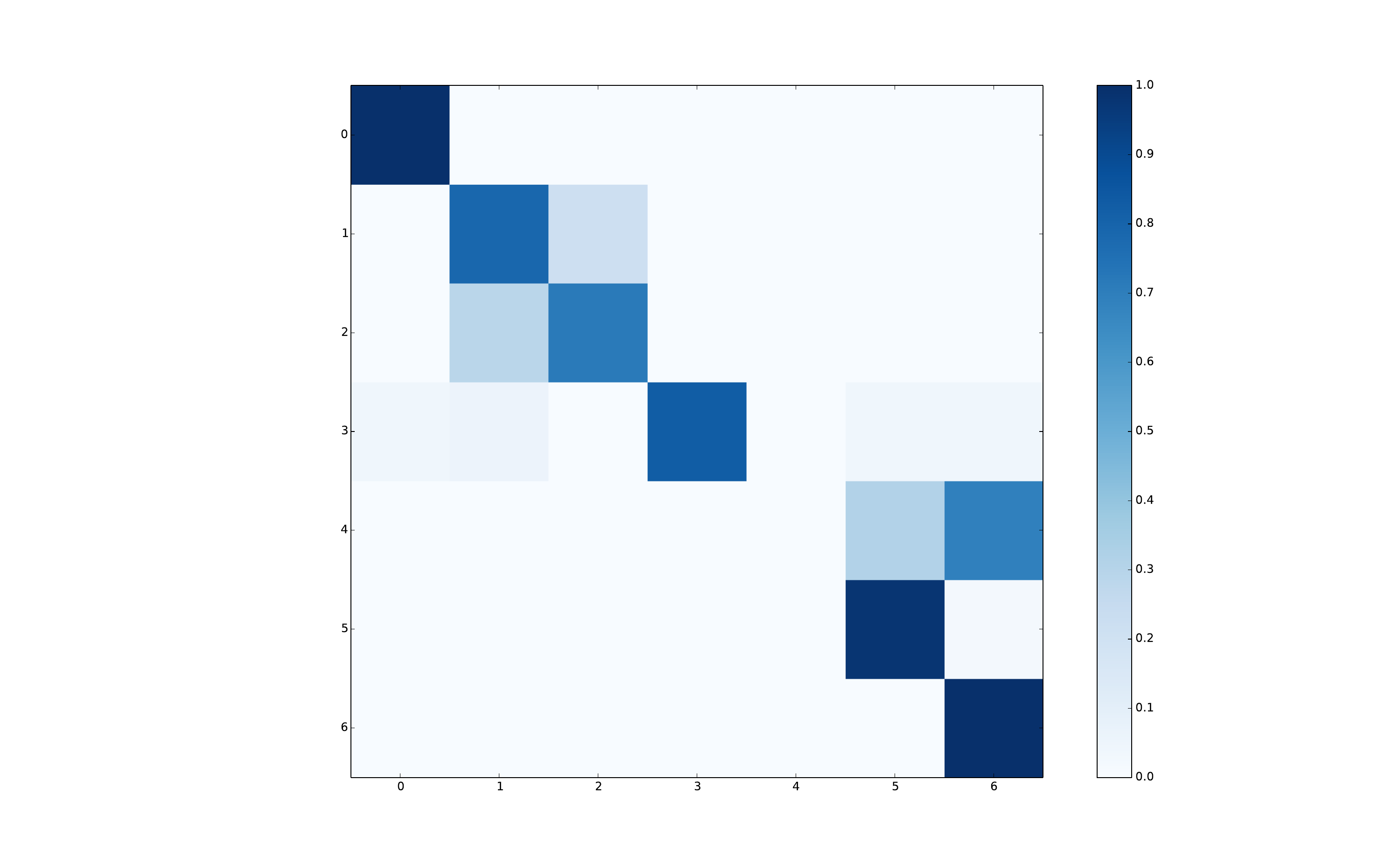}
        \caption{SBHAR}
        \label{sbharResult}
    \end{subfigure}
    \caption{Confusion Matrices of Proposed Method on PAMAP2, REALDISP, and SBHAR.}\label{confusionResult}
\end{figure*}

To evaluate the proposed network, three publicly available benchmark datasets, which contain wearable sensor data of different human activities, are used in the experiments to verify the effectiveness of the approach. The architectures of the models for the three datasets are manually chosen, \madd{including} \mdel{such as} the number of layers and the number of neurons in each layer. The architecture information is summarized in Table \ref{architecutreChoice}. The activation function used in the model is LeakyReLu.

\begin{table}[t]
\centering
\caption{The architecture of our approach for different datasets. Here, only the architecture of the encoder is shown. The decoder reverses the encoder.}\label{architecutreChoice}\smallskip
\scalebox{0.9}{%
\begin{tabular}{c|c}
\hline
\textbf{Dataset} &  \textbf{The number of neurons in each layer} \\
\hline
PAMAP2 & Input - 128 - 64 \\
\hline
REALDISP & Input - 256 - 128 \\
\hline
SBHAR  & Input - 30 - 20 \\
\hline
\end{tabular}}
\end{table}

\subsection{Datasets}
The three HAR datasets \add{used here are} \del{include} PAMAP2 \cite{pamap2}, REALDISP \cite{realdisp}, and SBHAR \cite{sbhar}. We use five-fold cross-validation to measure performance \del{. A} \add{a}nd all the sensor data sequences are segmented with the sliding window method.

PAMAP2 is collected from 9 participants performing 12 activities using 3 inertial measurement units placed on the wrist, chest and ankle. The dataset contains data of sport exercises (rope jumping, nordic walking\madd{,} etc.), and household activities (vacuum cleaning, ironing\madd{,} etc.). During the experiments, heart rate, accelerometer, gyroscope, magnetometer, and temperature data is recorded. \add{In accordance with previous research on this dataset, a}\del{A} sliding window of 5.12 seconds with one second step size is used to segment the data.

REALDISP is recorded from 17 volunteers carrying out 33 activities using 9 sensors placed on both arms, both legs, and the back. Each sensor provides acceleration, gyroscope and magnetic field orientation\del{ and quaternions}. This dataset contains data of fitness exercises, warm up, and cool down. The sliding window used here has a size of 2 seconds without overlapping.

SBHAR is gathered from 30 participants performing 6 basic activities, such as walking, lying, \madd{etc, } and 6 postural transitions, such as sit-to-lie\mdel{,} \madd{and} sit-to-stand. In our experiments, all the postural transitions are treated as one general transition. The dataset was collected \mdel{by} using a smartphone placed on the waist of the participants. The data is segmented using a sliding window of 2.56 seconds with a step size of 1.28 seconds.

\com{You need to say something about how you selected the cluster numbers. Did you select exactly the number of classes ? What would happen if you would give it more clusters than the number of classes ? Would they get cleaner ?  Also, you nowhere talked about your choices for $\alpha$ and $\beta$ and how you derived them. You also did not say anything regarding the size and architecture of your AE network.}

\subsection{Validation Metrics}
Three evaluation metrics are adopted to measure the performance of the approach: clustering accuracy (ACC), adjusted Rand index (ARI), and normalized mutual information (NMI). The ARI and NMI are computed as follows:
\[
\text{ARI} = \frac{\sum_{ij}\binom{n_{ij}}{2} -[\sum_{i}\binom{n_i}{2}\sum_{j}\binom{n_j}{2}] / \binom{n}{2}} {\frac{1}{2}[\sum_{i}\binom{n_i}{2} + \sum_{j}\binom{n_j}{2}] - [\sum_{i}\binom{n_i}{2}\sum_{j}\binom{n_j}{2}] / \binom{n}{2}}
\]

\[
\text{NMI} = \frac{\sum_i\sum_j n_{ij} \text{log}(\frac{n \cdot n_{ij}} {n_i \cdot n_j} ) } {\sqrt{\sum_i n_i \text{log}\frac{n_i}{n} \sum_j n_j \text{log} \frac{n_j}{n} }}
\]
where $n_{ij}$ is the number of samples in cluster $i$ and class $j$, $n_i$ is the number of samples in cluster $i$ \add{formed using the unsupervised appraoch}, $n_j$ is the number of samples in class $j$ \add{as indicated by the labels in the dataset}, and $n$ is the number of samples. 

\com{Would it be better to use different symbols for numbers of elements in clusters vs classes ? You are also not indicating here where the class labels come from. Also, you are not describing Clustering Accuracy here.}
\com{Should we include a reference as to what precision can be achieved using supervised learning ?}

\subsection{Results and Analysis}
\begin{table}[h!]
  \begin{center}
    \caption{The comparison between the proposed unsupervised approach and other unsupervised methods on the wearable sensor-based human activity datasets.}
    \scalebox{0.85}{%
    \label{unsupervisedResults}
    \begin{tabular}{c|c|c|c} % <-- Alignments: 1st column left, 2nd middle and 3rd right, with vertical lines in between
      \textbf{Methods} & \textbf{ACC} & \textbf{ARI}& \textbf{NMI} \\
      \hline
      \multicolumn{4}{c}{\textbf{PAMAP2}} \\
      \hline
%      KM & 0.6897 & 0.6360 & 0.7847 \\
      PCA + KM & $0.6993   $ & $0.6440  $ & $0.7905  $ \\
      AE + KM & $0.7706  $ & $0.6862  $ & $0.7994  $ \\
      Proposed Method + KM ($tn$)&$ 0.8543 $ & $0.8016$& $0.8730 $ \\
      Proposed Method + KM ($tn+1$)&$ 0.8622$ & $0.8089 $& $0.8898 $ \\
      Proposed Method + KM ($tn+2$)&$ \textbf{0.9211}$ & $0.8288 $& $0.8909 $ \\
      Proposed Method + KM ($tn+3$)&$ 0.9150$ & $\textbf{0.8590} $& $\textbf{0.9144} $ \\

      \hline
      \multicolumn{4}{c}{\textbf{REALDISP}} \\
      \hline
%      KM & 0.52591 & 0.37381& 0.6890 \\
      PCA + KM & $0.5723  $ & $0.4035  $ & $0.6890  $ \\
      AE + KM & $0.6401 $ & $0.5446  $ & $0.7764 $ \\
      Proposed Method + KM ($tn$)& $0.6812  $& $0.6051$ & $0.8043 $ \\
      Proposed Method + KM ($tn+1$)& $0.6829  $& $0.6062 $ & $0.7965  $ \\
      Proposed Method + KM ($tn+2$)& $0.7057 $& $0.6349 $ & $0.8215  $ \\
      Proposed Method + KM ($tn+3$)&$ \textbf{0.7149}$ & $\textbf{0.6508} $& $\textbf{0.8282} $ \\

      \hline
      \multicolumn{4}{c}{\textbf{SBHAR}} \\
      \hline
%      KM & 0.6579 & 0.5779 & 0.7188 \\
      PCA + KM & $0.6589 $ & $0.5784 $ & $ 0.7194  $ \\
      AE + KM & $ 0.6369  $ & $0.5090  $ & $0.7048  $ \\
      Proposed Method + KM ($tn$)& $ 0.7401 $ & $0.6343$ & $0.7569 $ \\
      Proposed Method + KM ($tn+1$)& $ 0.7596 $ & $\textbf{0.6718}  $ & $ \textbf{0.7982} $ \\
      Proposed Method + KM ($tn+2$)& $0.8018 $ & $0.6548 $ & $0.7552 $ \\
      Proposed Method + KM ($tn+3$)&$ \textbf{0.8073}$ & $ 0.6645 $& $ 0.7652 $ \\

    \end{tabular}}
  \end{center}
\end{table}

\begin{table}[h!]
  \begin{center}
    \caption{The comparison between the proposed unsupervised approach and other supervised methods on the wearable sensor-based human activity datasets.}
    \scalebox{0.85}{%
    \label{supervisedResults}
    \begin{tabular}{c|c} % <-- Alignments: 1st column left, 2nd middle and 3rd right, with vertical lines in between
      \textbf{Methods} & \textbf{ACC}  \\
      \hline
      \multicolumn{2}{c}{\textbf{PAMAP2}} \\
      \hline
%      KM & 0.6897 & 0.6360 & 0.7847 \\
      Probability SVM with Filter \cite{transitionAware} & $0.9304$    \\
      Decision Tree (C4.5) \cite{pamap2benchmark} & $0.9709$ \\
      Boosted C4.5 \cite{pamap2benchmark} & $\textbf{0.9980}$ \\
      Proposed Method + KM ($tn+2$)&$0.9211$  \\
      \hline
      \multicolumn{2}{c}{\textbf{REALDISP}} \\
      \hline
%      KM & 0.52591 & 0.37381& 0.6890 \\
      Probability SVM with Filter \cite{transitionAware} & $ \textbf{0.9952} $  \\
      kNN \cite{realdisp} & $0.9600 $   \\
      Proposed Method + KM ($tn+3$)& $0.7149 $ \\
      \hline
      \multicolumn{2}{c}{\textbf{SBHAR}} \\
      \hline
%      KM & 0.6579 & 0.5779 & 0.7188 \\
      Probability SVM with Filter \cite{transitionAware} & $0.9678$ \\
      CNN \cite{lundHAR} & $ \textbf{0.9870} $  \\
      Proposed Method + KM ($tn+3$)& $0.8073$ \\
    \end{tabular}}
  \end{center}
\end{table}

Our proposed approach is a\del{n} domain-specific extension based on traditional AE, so we compare the performance of the proposed method with the traditional AE and principal components analysis (PCA). The results of the experiments are summarized in Table \ref{unsupervisedResults}. 

As shown in the table, our approach achieves improved results over all three datasets. KM is applied in the embedding space. The number of clusters is chosen manually, and different cluster numbers are tested. The true number of classes in each dataset is used as the basis: $tn$. The results \del{proves}\add{show} that our approach can derive meaningful features to the subsequent activity clustering tasks. Figure. \ref{confusionAE} and \ref{confusionResult} also show and compare the performance of the traditional AE and the proposed approach by means of confusion matrices. 

In addition, Table \ref{supervisedResults} shows the comparison between the best results from our unsupervised approach and the results from previous published supervised methods on these three datasets. Note that, because ARI and NMI are metrics used to measure the performance of clustering algorithms, only ACC is adopted here to compare the results. As shown in the table, supervised methods can still achieve much better performance than unsupervised methods, but, as discussed before, the labeled data is usually difficult to acquire.

The results of the experiments \del{have} show\del{n} the efficiency of the approach, but we also noticed some inaccuracies introduced by this approach. One problem is locality preserving loss will mix some similar activities. For example, the activities jogging and running are located closely in the embedding space. The possible reason is that the difference between jogging and running is subtle. Jogging can be seen as a slow form of running. Moreover, different people jog or run at different speed\add{s}. Thus, the locality preserving loss can drive the model to project these two different activities to \del{the} adjacent locations in the embedding space. Another problem is \add{that} when a person switches between different activities, the temporal coherence assumption does not hold. During the activity transition process, the temporally adjacent data samples may represent different types of activities, hence the temporal coherence assumption will introduce inaccuracy into the model.

However, the problems mentioned above are usually infrequent. Therefore, the proposed approach can still learn useful representations and boost the performance of the subsequent clustering algorithm.
\com{Would it be useful to try an ablation study using different tradeoff settings ? Also, would it be useful to try it with more clusters and then combine the ones belonging to the same activities ?}

\subsection{Ablation Studies}
To further understand the effect of the different loss terms, a set of ablation experiments are conducted to measure the influences of the different loss terms separately. In the ablation experiments, the original joint loss function is transformed into two separate objectives: (i) the temporal coherence (TC) loss with the AE reconstruction loss; (ii) the locality preservation (LP) loss with the AE reconstruction loss. Each objective is used to train the model separately, thus comparing their results should provide insight into the benefit of each loss term. The results of the ablation studies are listed in Table \ref{ablation}.

\begin{table}[h!]
  \begin{center}
    \caption{Ablation studies on the effect of each loss term.}
    \scalebox{0.9}{%
    \label{ablation}
    \begin{tabular}{c|c|c|c} % <-- Alignments: 1st column left, 2nd middle and 3rd right, with vertical lines in between
      \textbf{Methods} & \textbf{ACC} & \textbf{ARI}& \textbf{NMI} \\
      \hline
      \multicolumn{4}{c}{\textbf{PAMAP2}} \\
      \hline
%      KM & 0.6897 & 0.6360 & 0.7847 \\
      TC Loss + AE Loss & $0.7859  $ & $0.7253 $ & $0.8140  $ \\
      LP Loss + AE Loss & $0.8065  $ & $0.7493  $ & $0.8337  $ \\
      Joint Loss &$ \textbf{0.8543}$ & $\textbf{0.8016} $& $\textbf{0.8730} $ \\
      \hline
      \multicolumn{4}{c}{\textbf{REALDISP}} \\
      \hline
%      KM & 0.52591 & 0.37381& 0.6890 \\
      TC Loss + AE Loss & $0.6341  $ & $0.5348  $ & $0.7639 $ \\
      LP Loss + AE Loss & $0.6610  $ & $0.5890  $ & $0.7829   $ \\
      Joint Loss & $\textbf{0.6812}  $& $\textbf{0.6051}  $ & $\textbf{0.8043}  $ \\
      \hline
      \multicolumn{4}{c}{\textbf{SBHAR}} \\
      \hline
%      KM & 0.6579 & 0.5779 & 0.7188 \\
      TC Loss + AE Loss & $0.6449 $ & $0.5173  $ & $ 0.7087  $ \\
      LP Loss + AE Loss & $ 0.7308  $ & $0.6182  $ & $0.7440  $ \\
      Joint Loss & $\textbf{0.7401} $ & $\textbf{0.6343}  $ & $\textbf{0.7569}  $ \\
    \end{tabular}}
  \end{center}
\end{table}
We notice that including both loss terms in the objective function can clearly improve the clustering performance. These results suggest that both loss terms guided the model to capture different useful information in the human activity sensor data in an unsupervised manner.

\section{Conclusion}
In this work, we have presented an unsupervised embedding learning approach, which is based on an autoencoder framework and uses the properties of human activities: temporal coherence and locality preserv\mdel{ing}\madd{ation}, to project the activity data into the embedding space. We have demonstrated the effectiveness of the approach by applying it to three widely used HAR benchmark datasets. The results of the experiments show that our approach can group similar activities together in the embedding space\mdel{,} \madd{and} therefore help improve the performance of the subsequent clustering task.

\bibliographystyle{aaai}
\bibliography{aaai2020}

\end{document}